\documentclass[11pt]{article}

\usepackage{acl}

\usepackage{times}
\usepackage{latexsym}

\usepackage[T1]{fontenc}

\usepackage[utf8]{inputenc}

\usepackage{microtype}

\usepackage{inconsolata}

\usepackage{graphicx}

\usepackage{url}
\usepackage{multirow}
\usepackage{booktabs}
\usepackage{amsmath}
\usepackage{color}
\usepackage{colortbl}
\usepackage{xurl}
\usepackage{CJKutf8}
\newcommand{\ja}[1]{\begin{CJK}{UTF8}{ipxm}#1\end{CJK}}

\newcommand{\blue}[1]{\textcolor{blue}{#1}}
\newcommand{\bblue}{\cellcolor[rgb]{.76,.90,.97}}
\newcommand{\bpink}{\cellcolor[rgb]{.98,.81,.86}}

\title{ATD-Trans: A Geographically Grounded Japanese--English Travelogue Translation Dataset}

\author{
Shohei Higashiyama${}^1$
\hspace{0.1cm}
Hiroki Ouchi${}^2$
\hspace{0.1cm}
Atsushi Fujita${}^1$
\hspace{0.1cm}
Masao Utiyama${}^1$\\
${}^1$National Institute of Information and Communications Technology\\
${}^2$Nara Institute of Science and Technology\\
\texttt{\{shohei.higashiyama,atsushi.fujita,mutiyama\}@nict.go.jp}\\
\texttt{hiroki.ouchi@is.naist.jp}
}

\begin{document}
\maketitle
\begin{abstract}
Geographic text, or textual data rich in geographic (geo-) information is a valuable source for various geographic applications, e.g., tourism management.
Making such information accessible to speakers of other languages further enhances its utility; thus, accurate machine translation (MT) is essential for equity in multilingual geo-information access.
To facilitate in-depth analysis for geographic text, we introduce ATD-Trans, a geographically grounded Japanese--English travelogue translation dataset, which enables evaluation of MT quality at both the overall and geo-entity levels across domestic (within Japan) and overseas regions.
Our experiments on existing language models examine two factors:  model language focus and geographic regions.
The results highlight advantages of Japanese-enhanced models and greater difficulty in translating domestic-region geo-entities mentioned in travel blogs.
\end{abstract}

\section{Introduction} \label{sec:intro}

Text can be regarded as an unstructured source of geographic information (geo-information),\footnote{Geo-information consists of a pair of spatial and non-spatial attribute information about a real-world entity~\cite{chirijohokagaku}. For example, the statement that ``an entity named Todaiji is located in Nara, Japan'' contains geo-information.} as it describes real-world entities from a human perspective. 
Textual data containing descriptions of geo-information (henceforth, \textit{geographic text data}) exhibit several notable characteristics:
(i) location information is often implicit, conveyed through geographic entity (geo-entity) mentions, such as place names and facility names, rather than explicit coordinates;
(ii) non-spatial factual and subjective information is diverse and rich; and
(iii) large volumes of web data exist, including user-generated contents.
Due to the characteristics (ii) and (iii), geographic text data is expected to support a wide range of applications, such as tourism management and disaster management~\cite{hu-etal-2022-location}. 

In multilingual information access, it is desirable that users of any language be able to obtain useful information from human- or machine-generated text, regardless of the language in which the text is written or the geographic region referred to in it. 
A key technology enabling this is machine translation (MT).
In practice, however, MT quality can vary across languages and geographic regions described in the text due to various factors, such as training data bias, potentially leading to inequities in information access. 
For example, if a non-Japanese speaker seeks information about a region in Japan described only in Japanese, low-quality MT texts translated into their language can place them at a disadvantage compared with speakers of Japanese or better-supported languages.

\begin{figure}[t] 
\centering
\includegraphics[width=7.7cm]{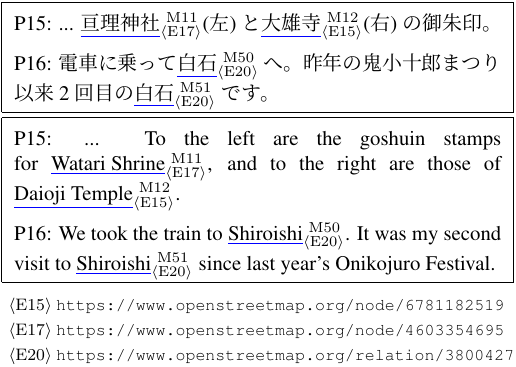}
\caption{A fragment of Japanese--English parallel text with geo-annotations from ATD-Trans (Document ID: 20851). Blue underlining indicates named geo-entity mentions. P, M, and E denote the prefixes of paragraph, mention, and entity IDs, respectively.
}
\label{fig:ex_doc}
\end{figure}

A central challenge in MT for geographic text is to accurately translate expressions that describe geo-entities, which can directly affect equity in geo-information access.
Additionally, the document context should be taken into account to resolve implicit or ambiguous geo-entity mentions and orthographic variants.
Many studies have investigated entity-aware MT for named entities (NEs)~\cite{conia-etal-2025-semeval}, as well as document-level MT for general documents~\cite{wang-etal-2023-document-level}. However, MT for geographic text remains underexplored, particularly with respect to differences in the geographic regions described in the text.

To facilitate in-depth analysis and thereby advance MT for geographic text, we constructed ATD-Trans, a geographically grounded Japanese--English translation dataset.
The dataset is derived from Japanese travel blogs about either domestic trips within Japan or overseas trips. As illustrated in Figure~\ref{fig:ex_doc}, the Japanese travel blog articles were translated into English, and both the Japanese and English articles were annotated with geo-entity mentions, along with their corresponding geographic knowledge base (KB) entry IDs.
This dataset enables document-level evaluation of (a) MT for texts rich in geo-entity mentions, which is the primary focus of this study, as well as (b) fundamental geographic text processing tasks, i.e., geo-entity mention recognition and geocoding.\footnote{While evaluation across translation directions is important, this study focuses on Japanese-to-English translation as a case study. Creating datasets for other translation directions using a similar data construction framework should enable evaluation for those directions as well.}

Using the created dataset, we have conducted MT experiments with existing large language models (LLMs) to evaluate translation quality at both the overall and geo-entity levels. 
We particularly focus on two factors: the language focus of the evaluated LLMs (English-centric vs.\ Japanese-enhanced) and the geographic region mentioned in the text (domestic vs.\ overseas).
From the experiments, we found that (1) Japanese-enhanced models generally outperformed their English-centric counterparts, suggesting better knowledge of geo-entities written in Japanese, and (2) texts about domestic trips were more difficult to translate, with geo-entity mentions accounting for a substantial part of the difficulty.
Furthermore, we found term-specifying prompts has both benefits and risks: it helps when it provides accurate knowledge that the model does not already possess, but it can degrade translation quality when the provided terms are inappropriate.%
\footnote{
The application form for access to ATD-Trans is available at \url{https://att-astrec.nict.go.jp/member/shigashiyama/resources/atd-trans/index.html}.
}

\section{Related Work}
\subsection*{MT of Lexically Constrained Expressions}
Accurate and consistent translation of lexically constrained expressions, such as NEs and terminology, has long been recognized as an important aspect of MT, and has increasingly been studied in the context of document-level MT in recent years~\cite{jin-etal-2023-challenges,wang-etal-2023-document-level}.
\citet{conia-etal-2025-semeval} organized the SemEval-2025 Entity-Aware MT shared task for texts with entities that require non-literal translation. 
Their results showed that current LLMs were not able to translate NEs perfectly even when provided with gold KB entry IDs, and that systems without such information yielded substantially lower translation quality.
\citet{semenov-etal-2025-findings} organized the WMT25 Terminology Translation Task. They showed that top systems achieved near-perfect terminology accuracy in the sentence-level track, whereas terminology translation remained more challenging in the document-level track, even when a glossary was provided.

While these studies examined how current MT systems handle general NEs and terminology, our study focuses on geographic text, paying particular attention to differences in the geographic regions mentioned in text.

\subsection*{Geographic and Cultural Information Access Across Languages}
Geoparsing~\cite{gritta-etal-2020-pragmatic,hu-etal-2022-location} is a fundamental task for extracting geo-information from text, aiming to identify geo-entity mentions and map them to geographic coordinates or to corresponding KB entries.\footnote{This mapping step is referred to as geocoding.}
A central challenge in geoparsing is the accurate analysis of fine-grained point-of-interest (POI)~\cite{nakatani-etal-2025-text}.
Previous studies \cite{gelernter2013,chen2015} have explored cross-lingual geoparsing, in which source texts are first machine-translated into a resource-rich language such as English and then the translated texts are geoparsed.
In both studies, the reported accuracy was comparable to that obtained by directly geoparsing the source text, although unfamiliar cites and POIs remained challenging. 
This cross-lingual strategy is useful for obtaining location information from text, particularly in low-resource languages.
However, from the perspective of equitable geo-information access, our focus differs in that we directly evaluate translation quality of MT outputs.

Another line of work has investigated cross-lingual accuracy differences of LLMs using multilingual parallel question answering benchmarks that require geographical and cultural knowledge: for example, \citet{singh-etal-2025-global} showed that LLM accuracy decreases in the order of high-, mid-, and low-resource languages, while \citet{roh-etal-2025-xlqa} demonstrated that LLMs perform poorly on locale-sensitive questions, in which the assumed geographic context and correct answer varies across languages, particularly for non-English languages.
These findings suggest that LLM knowledge varies across languages and regions, potentially leading to unequal information access.
Our study instead focuses on evaluating whether geo-information in the source text is preserved through MT.

\section{Dataset Construction}
To construct ATD-Trans, we used two existing travelogue datasets: 
the NAIST Academic Travelogue Dataset (ATD)\footnote{\url{https://sites.google.com/view/geography-and-language/resources}}~\cite{ouchi-etal-2023-atd} and ATD-MCL\footnote{\url{https://github.com/naist-nlp/atd-mcl}}~\cite{higashiyama-etal-2024-arukikata}.
ATD is a collection of Japanese travel blog posts published on a travel blogging platform, covering both domestic trips within Japan and overseas trips.
ATD-MCL is a subset of ATD consisting of domestic travel blogs annotated with three types of information: (i) geo-entity mentions, (ii) coreference relations among the mentions, and (iii) links to corresponding entries in OpenStreetMap (OSM).\footnote{\url{https://www.openstreetmap.org/}}

In this study, we constructed 
(a) a geographically annotated dataset of overseas travel blogs from ATD, annotated with the same types of information as ATD-MCL;
(b) an English translation dataset of domestic and overseas travel blogs from ATD; and
(c) ATD-Trans, the unified dataset integrating these two resources.

Table~\ref{tab:atd_stat_overview} summarizes the relationships among ATD and its derivatives.
The proposed dataset, ATD-Trans, consists of a total of 90 translated articles together with their corresponding source articles and geo-annotations.

Figure~\ref{fig:ex_doc} illustrates a fragment of parallel text from the ATD-Trans dev set.
The mention ``\ja{白石}'' is ambiguous because multiple locations in Japan share the reading \textit{Shiraishi} or \textit{Shiroishi}.
However, context both within and outside the paragraph, such as ``Watari Shrine'' and ``Onikojuro Festival,'' makes it possible to disambiguate the mention as referring to \textit{Shiroishi}, Miyagi Prefecture.

\begin{table}[t]
\small
\centering
\begin{tabular}{l|ccc}
\toprule
& Source & Geo-Anno & Target \\
\midrule
Domestic articles & 4,500 & 200 & \underline{58} \\
Overseas articles & 9,500 &  \underline{78} & \underline{32} \\
\bottomrule
\end{tabular}
\caption{Number of articles in the original ATD and its derivatives.
Source and target denote the original Japanese travel blogs and their English translations, respectively, and geo-anno denotes geographic annotations on source/target articles.
ATD consists of 14,000 source articles, and ATD-MCL corresponds to the 200 domestic source articles with geo-annotation.
}
\label{tab:atd_stat_overview}
\end{table}

\subsection{Geo-Annotations of Overseas Articles}
Following the annotation guidelines and procedures for ATD-MCL, two annotators annotated mentions and coreference relations, and five annotators assigned OSM links.
All annotators were native Japanese speakers employed by an external annotation company.
The dataset comprises 78 articles, 4,313 sentences, and 5,116 mentions.

To assess annotation quality, two annotators independently annotated mentions and coreference relations in five of the 78 articles, and annotated OSM links in three of those five articles.
Inter-annotator agreement was as follows: F1 score of 0.90 for mentions, LEA score~\cite{moosavi-strube-2016-coreference} of 0.86 for coreference, and Cohen’s $\kappa$ of 0.69 for OSM links, indicating high agreement.

\subsection{English Translation}
The translation work was conducted by a professional translation company in accordance with the workflow specified in ISO 17100:2015~\cite{iso17100-2015}, and involved two native Japanese translators specializing in Japanese--English translation and three native English bilingual checkers.
A total of 90 articles were translated.

The translation specifications were as follows:
(1) Each article was translated by a single translator at the paragraph level, ensuring that the resulting English text is natural and well structured.
(2) In the Japanese source text, geo-entity mentions were enclosed in predefined tags, and the corresponding expressions in the English translation were required to be enclosed with the same tags.
(3) For geo-entities and other NEs, established and commonly used English names were adopted when available. 
If the same NE appeared multiple times within an article, translators were permitted to use pronouns or omit repeated mentions where appropriate.

\section{Experimental Settings}
We report MT experiments using our dataset to evaluate existing LLMs. 
As stated in \S\ref{sec:intro}, the primary objective is to analyze performance trends with respect to the language focus of the LLMs (English-centric vs.\ Japanese-enhanced) and geographic region mentioned in the text (Japan vs.\ overseas).
We further examine the effects of document-level context and explicit term specification by grounding mentions in a geographic KB.

\paragraph{Experimental Dataset.}
To balance data size between domestic and overseas travel blogs, we selected 32 out of the 58 domestic articles and used all the 32 overseas articles. Each subset was divided into 15, 6, and 11 articles, corresponding to the train, dev, and test sets, respectively. Descriptive statistics of the dataset are shown in Table~\ref{tab:data_split}.\footnote{Although the dataset was also annotated with nominal mentions such as pronouns, the ``\#Mentions'' values in Table~\ref{tab:data_split} indicates the number of named mentions. The same applies to the Term Accuracy evaluation and the experiments with term specifying prompts described later.}

\begin{table}[t]
\footnotesize
\centering
\begin{tabular}{llccc}
\toprule
&& \multicolumn{1}{c}{\#Paras} & \multicolumn{1}{c}{\#Sentences} & \multicolumn{1}{c}{\#Mentions} \\
\midrule
& Train & 421 & 1,120\,/\,1,100 & 629\,/\,627 \\
Dom & Dev   & 199 &   440\,/\,434   & 231\,/\,230 \\
& Test  & 330 & \!\!\!\!1,023\,/\,971 & 493\,/\,493 \\
\midrule
& Train & 238 &   869\,/\,919 & 552\,/\,554 \\
Ovs & Dev   & 131 &   409\,/\,414 & 189\,/\,189 \\
& Test  & 190 &   611\,/\,689 & 361\,/\,360 \\
\bottomrule
\end{tabular}
\caption{Statistics of experimental dataset (Ja/En).}
\label{tab:data_split}
\end{table}

\paragraph{Evaluated Models.}
Among existing autoregressive LLMs, we evaluated eight instruction-tuned models:
 Llama 3.1 (8B, 70B)~\cite{grattafiori2024llama3}, Gemma 2 (9B, 27B)~\cite{gemmateam2024gemma2improvingopen}, Llama-3.1-Swallow (8B, 70B) and Gemma-2-Llama Swallow (9B, 27B)~\cite{Fujii-COLM2024}.
Llama 3.1 and Gemma 2 are English-centric models, whereas Llama-3.1-Swallow and Gemma-2-Llama Swallow are Japanese-enhanced models derived from the corresponding base models (i.e., Llama 3.1 base or Gemma 2 base).
The two Swallow model series, Llama-3.1-Swallow and Gemma-2-Llama Swallow, were trained using similar corpora through continual pretraining and post-training; however, the newer Gemma-2-Llama Swallow models incorporate more recent techniques, such as corpus filtering and LLM-synthesized data, for creating the training data.

\paragraph{Inference Settings.}
We used publicly available checkpoints of the above models and evaluated their outputs without additional fine-tuning, performing zero-shot in-context learning under several inference and prompting conditions.
We conducted paragraph-level inference under two context settings: (1) using only the target paragraph as context and (2) using the entire document as context to generate the translation of the target paragraph.\footnote{We conducted preliminary experiments on document-level inference with instructions to preserve paragraph boundaries. However, some models did not preserve them, and translation quality across models was comparable to that of paragraph-level inference with document context. We therefore excluded document-level inference from our experiments.}
Furthermore, we examined several prompting conditions for term specification.
These settings are summarized below, and the full prompt templates are provided in Appendix~\ref{app:exp_settings}:
\begin{itemize}
\setlength{\parskip}{0cm} 
\setlength{\itemsep}{0.1cm}
\item a basic translation prompt with paragraph context  (Context=``Para'', Term=``None'');
\item a basic translation prompt with document context (Context=``Doc'', Term=``None'');
\item a document-context prompt specifying bilingual term pairs derived from the OSM entry names of the predicted or gold entry (Context=``Doc'', Term=``KB pred'' or ``KB gold'');
\item a document-context prompt specifying the oracle terms extracted from the reference translation (Context=``Doc'', Term=``Oracle'').
\end{itemize}
For the ``KB pred'' term prompts, we conducted experiments only on the domestic data, using POI Geocoder and its accompanying database file\footnote{\url{https://github.com/naist-nlp/poi-geocoding}}
\cite{nakatani-etal-2025-text} to predict OSM entries.\footnote{We did not conduct experiments on the overseas data because no preprocessed databases of entries outside Japan were available.} Details of the geocoder settings are provided in Appendix~\ref{app:geocoding}.
For each model, we set the temperature to 0.2 and report the average score over two runs with different random seeds.

\begin{table*}[t]
\small
\centering
\begin{tabular}{lc|ccc|ccc}
\toprule
\multirow{2}{*}{System} & \multirow{2}{*}{Context} & \multicolumn{3}{c|}{Domestic data} & \multicolumn{3}{c}{Overseas data} \\
\cmidrule{3-8}
&& \texttt{d-BLEU} & \texttt{COMET} & \texttt{TermAcc} & \texttt{d-BLEU} & \texttt{COMET} & \texttt{TermAcc} \\
\midrule
\multirow{2}{*}{Llama-3.1-8b-Instruct} 
& Para & 22.5 & 77.6 & \underline{41.6} & 29.6 & 82.9 & 68.7 \\
& Doc  & \underline{24.0} & \underline{78.1} & 38.6 & \underline{31.9} & \underline{83.9} & \underline{73.2} \\
\midrule
\multirow{2}{*}{Llama-3.1-Swallow-8B-Instruct-v0.3} 
& Para & 28.6 & 79.0 & \underline{51.4} & 33.1 & 84.2 & 73.5 \\
& Doc  & 29.5 & 79.0 & 48.0 & 33.2 & 83.9 & \underline{76.8} \\
\midrule
\multirow{2}{*}{Llama-3.1-70b-Instruct}
& Para & 31.5 & 79.9 & 59.5 & 35.7 & 84.2 & 78.3 \\
& Doc  & 32.1 & \underline{81.0} & 59.1 & 36.5 & \underline{\textbf{85.3}} & \underline{82.7} \\
\midrule
\multirow{2}{*}{Llama-3.1-Swallow-70B-Instruct-v0.3}
& Para & 33.2 & 80.7 & 61.5 & 38.0 & \textbf{85.3} & 80.8 \\
& Doc  & \textbf{33.8} & \underline{\textbf{81.9}} & \textbf{61.6} & \textbf{38.4} & 85.2 & \underline{\textbf{84.1}} \\
\midrule
\midrule
\multirow{2}{*}{gemma-2-9b-it}
& Para & 29.7 & 79.7 & \underline{50.1} & 33.9 & 84.3 & 74.3 \\
& Doc  & 29.8 & 79.6 & 45.2 & \underline{35.2} & 85.0 & \underline{79.0} \\
\midrule
\multirow{2}{*}{Gemma-2-Llama-Swallow-9b-it-v0.1}
& Para & 30.7 & 79.5 & 56.3 & 34.7 & 83.9 & 76.3 \\
& Doc  & 30.6 & 79.6 & 55.5 & 34.5 & 83.9 & \underline{78.5} \\
\midrule
\multirow{2}{*}{gemma-2-27b-it}
& Para & 31.4 & 79.8 & 57.2 & 36.6 & 84.5 & 78.8 \\
& Doc  & \underline{32.7} & 80.7 & 57.5 & 36.4 & 85.2 & \underline{79.9} \\
\midrule
\multirow{2}{*}{Gemma-2-Llama-Swallow-27b-it-v0.1}
& Para & 32.1 & 80.2 & 60.0 & 36.6 & 84.7 & 80.5 \\
& Doc  & \underline{33.2} & \underline{81.2} & 59.3 & 36.6 & 85.0 & 81.1 \\
\bottomrule
\end{tabular}
\caption{Translation quality of LLMs with basic prompts. For each model, the score that is higher by 1 point or more between the two context settings (Para: paragraph context; Doc: document context) is \underline{underlined}.} \label{tab:res_mt1}
\end{table*}

\paragraph{Evaluation Metrics.}
To measure overall translation quality, we used d-BLEU~\cite{liu-mbart}
and COMET (wmt22-comet-da)~\cite{rei-etal-2022-comet}.
To measure entity translation quality, we used the Term Success Ratio (TSR)~\cite{semenov-etal-2023-findings}, which measures whether the system output includes the reference translation of each geo-entity expression in the source text, using fuzzy matching with a threshold of 0.9. 
For clarity, we refer to this metric as Term Accuracy throughout this paper.

\section{Results and Analysis}
\subsection{Results for Basic Prompting} \label{sec:overall_result}
To understand the general performance trends of the models, we compared translation quality across four factors: context setting, model size, language focus of the models, and the geographic region mainly mentioned in the text (domestic vs.\ overseas).
Table~\ref{tab:res_mt1} presents results on the test sets.

Across the two context settings, document context generally yielded comparable or slightly higher d-BLEU and COMET scores across many models on both datasets. 
Document context consistently improved Term Accuracy for all models on the overseas data, whereas a slight decrease was observed for many models on the domestic data. 
These results suggest that using document context can improve overall translation quality, but it does not always lead to better term translation quality.

Within each of the four model series, larger models with the same context consistently achieved higher scores than smaller ones. The improvement was particularly notable for Term Accuracy. For example, Term Accuracy increased by about 9--20 points for Llama-3.1 (8B vs.\ 70B) and by about 7--14 points for Llama-3.1-Swallow (8B vs.\ 70B). 

\begin{table*}[t]
\small
\centering
\begin{tabular}{lc|ccc|ccc}
\toprule
\multirow{2}{*}{System} & \multirow{2}{*}{Entities} & \multicolumn{3}{c|}{Domestic data} & \multicolumn{3}{c}{Overseas data} \\
\cmidrule{3-8}
&& \texttt{d-BLEU} & \texttt{COMET} & \texttt{TermAcc}${}_1$ & \texttt{d-BLEU} & \texttt{COMET} & \texttt{TermAcc}${}_1$ \\
\midrule
\multirow{2}{*}{Llama-3.1-8b-Instruct} 
& Original & 24.0 & 78.1 & 35.8 & 31.9 & 83.9 & 69.5 \\
& ID-replaced & 27.4 & 80.8 & 96.7 & 29.2 & 82.9 & 96.0 \\
\midrule
\multirow{2}{*}{Llama-3.1-Swallow-8B-Instruct-v0.3} 
& Original & 29.5 & 79.0 & 43.0 & 33.2 & 83.9 & 72.5 \\
& ID-replaced & 29.6 & 81.1 & 96.4 & 32.0 & 82.8 & 96.1 \\
\midrule
\multirow{2}{*}{gemma-2-9b-it}
& Original & 29.8 & 79.6 & 40.0 & 35.2 & 85.0 & 74.6 \\
& ID-replaced & 30.0 & 81.8 & 98.1 & 33.3 & 84.2 & 98.1 \\
\midrule
\multirow{2}{*}{Gemma-2-Llama-Swallow-9b-it-v0.1}
& Original & 30.6 & 79.6 & 47.3 & 34.5 & 83.9 & 74.8 \\
& ID-replaced & 30.7& 81.7 & 97.3 & 33.0 & 83.2 & 97.5 \\
\bottomrule
\end{tabular}
\caption{Translation quality of LLMs on ID-replaced data.} \label{tab:res_mt_idrep}
\end{table*}

Compared with English-centric models, their corresponding Japanese-enhanced counterparts achieved comparable or better translation quality in most model--context combinations. 
Specifically, Term Accuracy improved for almost all pairs of English-centric and Japanese-enhanced models under the same context setting.
d-BLEU and COMET showed a more complex pattern: improvements were observed for both the Llama-3.1/Llama-3.1-Swallow pairs (8B and 70B) and gemma-2/Gemma-2-Llama-Swallow pairs (9B and 27B) on the domestic dataset, whereas on the overseas dataset, gains were obserbed for the former pairs but were limited for the latter pairs.
These results suggest that continual pretraining on Japanese data\footnote{Monolingual texts and Ja--En parallel texts.} enhances knowledge of geo-entities in Japan, including their English expressions, and also improves Japanese-to-English translation more broadly.

Across the domestic and overseas data under the same model and context setting, the overseas data consistently achieved higher scores. Overall, the overseas data yielded approximately 3--8 points higher d-BLEU, 4--6 points higher COMET, and 19--35 points higher Term Accuracy than the domestic data. 
Possible explanations include the following: overseas travel blogs may contain sentences that are generally easier to translate, or geo-entities appearing in overseas travel blogs may be easier to translate. Regarding the latter possibility, both domestic and overseas travel blogs in our dataset were primarily written in Japanese by authors who were likely residents of Japan. Overseas travel blogs may therefore contain geo-entities that are more widely known and less locally specific.

Overall, the results showed several expected trends: document context tended to improve translation quality, larger models outperformed their smaller counterparts, and Japanese-enhanced models outperformed their English-centric counterparts in many cases. 
In contrast, the higher scores observed for the overseas data require further analysis; we investigate this issue in detail in \S\ref{sec:syn_ids}.

\subsection{Analysis with Synthetic Geo-Entity IDs} \label{sec:syn_ids}
To test the above hypothesis that sentences or geo-entities in overseas travel blogs are easier to translate, we created synthetic domestic and overseas datasets by replacing geo-entity mentions in the original texts with alphanumeric identifiers, such as \texttt{L001}, while preserving the alignments between source and reference texts.

\begin{table*}[t]
\small
\centering
\begin{tabular}{lcc|ccc|ccc}
\toprule
\multirow{2}{*}{System} & \multirow{2}{*}{Context} & \multirow{2}{*}{Term} & \multicolumn{3}{c|}{Domestic data} & \multicolumn{3}{c}{Overseas data} \\
\cmidrule{4-9}
&&& \texttt{\!d-BLEU\!} & \texttt{COMET} & \texttt{\!TermAcc\!} & \texttt{\!d-BLEU\!} & \texttt{COMET} & \texttt{\!TermAcc\!} \\
\midrule
\multirow{4}{*}{Llama-3.1-8b-Instruct} 
& Doc & None & 24.0 & 78.1 & 38.6 & 31.9 & 83.9 & 73.2 \\
& Doc & KB pred & \bpink 20.4 & \bpink 77.1 & \bblue 46.9 & -- & -- & -- \\
& Doc & KB gold & \bpink 20.3 & \bpink 77.0 & \bblue 46.2 & \bpink 25.2 & \bpink 82.3 & \bblue 74.9 \\
& Doc & Oracle & \bblue 28.3 & \bblue 80.5 & \bblue 97.9 & 31.6 & 83.3 & \bblue 97.8 \\
\midrule
\multirow{4}{*}{Llama-3.1-Swallow-8B-Instruct-v0.3\!\!\!} 
& Doc & None & 29.5 & 79.0 & 48.0 & 33.2 & 83.9 & 76.8 \\
& Doc & KB pred & 29.2 & 78.7 & \bblue 51.5 & -- & -- & -- \\
& Doc & KB gold & 29.2 & 78.7 & \bblue 51.4 & \bpink 32.0 & 83.0 & 76.8 \\
& Doc & Oracle & \bblue 34.7 & \bblue 81.5 & \bblue 99.4 & 33.5 & 84.1 & \bblue 98.5 \\
\midrule
\multirow{4}{*}{Llama-3.1-70b-Instruct}
& Doc & None & 32.1 & 81.0 & 59.1 & 36.5 & 85.3 & 82.7 \\
& Doc & KB pred & \bpink 28.7 & \bpink 77.9 & \bpink 54.6  & -- & -- & -- \\
& Doc & KB gold & \bpink 28.7 & \bpink 77.8 & 58.7 & \bpink 32.8 & \bpink 81.6 & \bpink 76.8 \\
& Doc & Oracle & \bblue 33.4 & 81.4 & \bblue 100 & \bpink 35.5 & \bpink 83.8 & \bblue 99.7 \\
\midrule
\multirow{4}{*}{Llama-3.1-Swallow-70B-Instruct-v0.3\!\!\!}
& Doc & None & 33.8 & 81.9 & 61.6 & 38.4 & 85.2 & 84.1 \\
& Doc & KB pred & \bpink 32.5 & \bpink 79.9 & \bpink 56.7 & -- & -- & -- \\
& Doc & KB gold & \bpink 32.5 & \bpink 79.8 & \bpink 56.8 & \bpink 35.3 & \bpink 83.4 & \bpink 76.7 \\
& Doc & Oracle & \bblue 37.0 & \bblue 82.4 & \bblue 99.6 & 37.7 & 84.5 & \bblue 99.1 \\
\bottomrule
\end{tabular}
\caption{Translation quality of Llama-3.1 and Llama-3.1-Swallow with term-specifying prompts. The score that is 1 point higher or lower than that of the basic prompt is highlighted in light blue or pink, respectively.} \label{tab:res_mt2}
\end{table*}

\begin{table*}[t]
\small
\centering
\begin{tabular}{l|lc|ccc|c|cc}
\toprule
&&& \multicolumn{4}{c|}{\texttt{TermAcc}} && \\
Data & System & Term & KB-Matched & KB-Matched & KB-Unmatched & Total & \texttt{d-BLEU} & \texttt{COMET} \\
&&& (Ref-Cons) & (Ref-Incons) &&&& \\
\midrule
\multirow{6}{*}{Domestic} 
& (Num.) &&  185 & 65 & 218 & 468 & -- & -- \\
\cmidrule{2-9}
& \multirow{2}{*}{Swallow-8B} 
& None & 0.665 & 0.285 & 0.381 & 0.480 & 29.5 & 79.0 \\
&& KB gold & \bblue 0.870 & \bpink 0.092 & \bpink 0.337 & \bblue 0.514 & 29.2 & 78.7 \\
\cmidrule{2-9}
& \multirow{2}{*}{Swallow-70B} 
& None & 0.822 & 0.369 & 0.516 & 0.616 & 33.8 & 81.9 \\
&& KB gold & \bblue 0.881 & \bpink 0.062 & \bpink 0.454 & \bpink 0.568 & \bpink 32.5 & \bpink 79.8 \\
\midrule
\multirow{6}{*}{Overseas} 
& (Num.) && 207& 59 & 57 & 323 & -- & -- \\
\cmidrule{2-9}
& \multirow{2}{*}{Swallow-8B} 
& None & 0.891 & 0.339 & 0.763 & 0.768 & 33.2 & 83.9 \\
&& KB gold & \bblue 0.957 & \bpink 0.169 & \bpink 0.702 & 0.768 & \bpink 32.0 & 83.0 \\
\cmidrule{2-9}
& \multirow{2}{*}{Swallow-70B} 
& None & 0.911 & 0.542 & 0.895 & 0.841 & 38.4 & 85.2 \\
&& KB gold & \bblue 0.971 & \bpink 0.102 & \bpink 0.825 & \bpink 0.786 & \bpink 35.3 & \bpink 83.4 \\
\bottomrule
\end{tabular}
\caption{Term Accuracy of the Llama-3.1-Swallow 8B and 70B models (Swallow-8B and Swallow-70B) using basic and KB gold term prompts with document-level context, under KB-matching conditions.} \label{tab:analysis_term_acc}
\end{table*}

Table~\ref{tab:res_mt_idrep} shows the results obtained with a prompt (Appendix~\ref{app:exp_settings}) similar to the basic prompt with document context, but additionally instructing the model to preserve ID mentions. Here, we measured exact-match Term Accuracy (\texttt{TermAcc}${}_1$) instead of fuzzy match to distinguish similar mentions like \texttt{L001} and \texttt{L002}.
For both datasets, Term Accuracy was close to 100\%, indicating that the ID mentions were translated almost perfectly. 
Compared with the original data, overall translation quality, as measured by d-BLEU and COMET, tended to improve on the domestic data but decline on the overseas data, resulting in a smaller gap between the domestic and overseas datasets.
These findings suggest that translation difficulty for the models is largely attributable to entity mentions, and that the entities in the domestic data are more difficult to translate. 
An explanation for the still lower translation quality on the domestic data after ID replacement is that the data may contain culturally dependent expressions other than the replaced geo-entity mentions.

\subsection{Results for Term-Specifying Prompting} \label{sec:term_prompt}
To evaluate the effect of term specification, we compared three term-specifying prompts with the basic prompt using document context. For brevity, Table~\ref{tab:res_mt2} shows only the results for the Llama-3.1 and Llama-3.1-Swallow model series; results for the other models are provided in the Appendix~\ref{app:gemma}.

First, with the oracle term prompt, Term Accuracy reached nearly 100\% for all model--dataset combinations. Compared with the basic prompt, the oracle term prompt consistently improved d-BLEU and COMET scores on the domestic data, while producing comparable or slightly lower scores on the overseas data in some cases. These results indicate that the models were able to incorporate the specified terms as instructed without substantially degrading overall translation quality.

Next, for the KB gold and KB pred term prompts, d-BLEU and COMET scores clearly decreased relative to the basic prompt in most model--dataset combinations. For Term Accuracy, improvements and degradations were observed with roughly equal frequency.
A possible reason for the degradation is the non-negligible number of mismatches between KB terms and reference terms, which we examine in more detail in \S\ref{sec:term_acc}. 
The small difference between the gold-term and predicted-term prompts appears to be mainly due to two factors, according to our manual analysis: (1) the geocoding model tended to be accurate for well-known or less ambiguous geo-entities; (2) for coarse-grained POIs, the corresponding entries often lacked an English-name attribute, so bilingual term information was often not included in the translation prompt.

\subsection{Analysis of KB-Aware Term Accuracy} \label{sec:term_acc}
To further analyze the limited effect of KB term prompting, we compared Term Accuracy for the Llama-3.1-Swallow models (8B and 70B) under the basic prompt (Term=``None'') and the KB gold term prompt across the following three group of cases: 
(i) cases in which a KB entry corresponding to the mention exists (KB-Matched) and its term, i.e., the English name, is consistent with the reference (Ref-Cons), (ii) KB-Matched cases in which the KB term is inconsistent with the reference (Ref-Incons),\footnote{Consistency was determined based on Term Accuracy.} and (iii) cases in which no corresponding KB entry exists (KB-Unmatched).

Table~\ref{tab:analysis_term_acc} presents the results.\footnote{In Table~\ref{tab:analysis_term_acc}, the number of mentions was counted by treating identical string mentions within the same paragraph as a single mention. In the term-specifying prompt experiments, term specification was performed using this unit. However, Term Accuracy was computed over all individual mentions.}
A substantial proportion of mentions were matched to KB entries, with KB-Matched (Ref-Cons and Ref-Incons) cases accounting for 53\% and 82\% of all cases in the domestic and overseas datasets, respectively. 

For KB-Matched cases, Term Accuracy increased by 0.06--0.20 points in the Ref-Cons subset, whereas it decreased by 0.17--0.44 points in the Ref-Incons subset. This result is intuitive: as long as the model follows the prompt faithfully, whether the term specified in the prompt matches the reference term directly affects whether Term Accuracy improves or degrades. 
For KB-Unmatched cases, Term Accuracy decreased by 0.05--0.07 points. Because no term information was provided for these mentions in the KB term prompt, little difference from the basic prompt would be expected. This result suggests that specifying terms for other mentions may also influence the translation of mentions without explicit term information.

Overall, the results suggest two main points. First, the 8B model showed lower Term Accuracy under the basic prompt, leaving more room for improvement in Ref-Cons cases and less room for degradation in Ref-Incons cases; the 70B model showed higher Term Accuracy and the opposite tendency. Second, under the KB gold term prompt, neither Term Accuracy nor overall translation quality showed substantial degradation for the 8B model, whereas larger degradation was observed for the 70B model. This suggests that the 70B model possesses richer knowledge of Japanese places, benefiting less from additional bilingual term information while being more adversely affected by noisy term information.
Taken together, these results suggest that term information is most beneficial when it provides accurate knowledge that the model does not already possess.

\subsection{Translation Examples} \label{sec:mt_example}
We discuss example model outputs for geo-entities.
The full source, reference, and output sentences are provided in Appendix~\ref{app:mt_example}.

In a dev sentence, ``\ja{美濃戸口}'' (read as \textit{Minoto}\textit{guchi}) is correctly translated by the Llama-3.1-Swallow 70B model, whereas it is mistranslated as \textit{Minamidobuchi} by the Llama-3.1 70B model, an entirely unrelated reading.
Taken together with similar observations in other examples, this suggests that the latter model sometimes lacks accurate knowledge of the correct readings of kanji-based local place names in Japan.

In another dev sentence, for ``\ja{外宮}'' of Ise Jingu (romanized form: \textit{Geku}; literal translation: \textit{Outer Shrine}), the Llama-3.1-Swallow 70B model with the KB gold term prompt used the specified term, ``Ise-Jingu Geku,'' which was close to ``Geku'' in the reference translation, whereas the model with the basic prompt produced ``Outer Shrine.''
This example indicates that incidental matches or mismatches between KB-specified terms and reference translations can influence automatic translation metric scores.
Thus, lower metric scores do not necessarily imply unacceptable translations; desirable geo-entity translations depend on translation purposes and specifications.
In practical scenarios, it is therefore necessary to decide whether to adopt KB-based term specification by assessing its compatibility with the intended translation style.

\subsection{Summary}
The main findings regarding our primary focuses are summarized as follows.
Regarding the language focus of the LLMs, Japanese-enhanced models generally outperformed their English-centric counterparts, suggesting that the former have richer and more accurate knowledge of geo-entities in Japan thanks to the continual pre-training on massive Japanese text data (\S\ref{sec:overall_result}).
Regarding the geographic region mentioned in the text, we found that the overseas data was easier to translate than the domestic data, largely because geo-entity mentions account for a substantial part of the translation difficulty and may be less locally specific than those in the domestic data (\S\ref{sec:overall_result}--\ref{sec:syn_ids}). 

We observed several additional trends consistent with our expectations: document context was generally helpful, larger models performed better (\S\ref{sec:overall_result}), and recent LLMs were able to follow term-specifying instructions under oracle term prompting (\S\ref{sec:term_prompt}). However, KB-based term prompts often degraded translation quality, likely because of mismatches between KB terms and reference terms and negative effects on KB-Unmatched cases (\S\ref{sec:term_acc}). These results suggest that term information is most useful when it provides accurate knowledge that the model does not already possess.
\section{Conclusion}
We presented ATD-Trans, a geo-annotated Japanese--English travelogue translation dataset.
Our experiments on existing LLMs revealed two main findings regarding the model language focus and geographic regions described in the text.
In addition, we identified a remaining challenge in acquiring accurate geographic knowledge from an external knowledge base.

Future directions include expanding the dataset to additional domains and translation directions, developing systems that better leverage geographic knowledge bases, and designing extrinsic tasks and evaluation metrics tailored to real-world geographic applications such as route planning, as discussed in the Limitations section.

\clearpage
\section*{Limitations}
\paragraph{Dataset Scope and Coverage.}
Our dataset is restricted in terms of text domains and geographic regions (Japanese travel blogs describing domestic trips within Japan and overseas trips), as well as translation directions (Japanese to English).
Extending the dataset to additional domains, geographic regions, and translation directions remains an important direction for future work.
Despite these limitations, our study provides initial insights as a first case study to evaluate MT of geographic text from a region-focused perspective.

\paragraph{System Design for KB-Augmented MT.}
Our experiments rely on simple prompt designs and a simple geo-entity linking approach, but more robust or improved performance may be achievable through extensive prompt engineering or additional system components, at the cost of increased computation and implementation complexity.
In particular, OpenStreetMap often contains multiple entries corresponding to similar or equivalent geographic features.
Future work includes incorporating more sophisticated selection mechanisms to identify entries that are appropriate with respect to both geographic location and multilingual name attributes, potentially by leveraging additional external knowledge sources such as Wikipedia.

\paragraph{Evaluation Beyond Intrinsic MT Quality.}
MT of geographic text is closely connected to real-world applications, such as supporting travel route planning and evacuation guidance.
In such scenarios, it is essential that users can correctly identify and reach intended routes or destinations through machine-translated text, which requires accurate translation of all geo-entity mentions and their relationships.
This highlights the need for extrinsic task design and evaluation criteria beyond intrinsic MT evaluation, such as success rate of extrinsic tasks, which remains an important direction for future work.

\section*{Ethics Statemets}
\paragraph{License of Resources}
ATD is available for research purposes under its terms of use.
ATD-MCL, POI Geocoder, spaCy, GiNZA, LUKE-NER are available under the MIT License.
Each language model checkpoint is available under its respective license.
Our use of these resources for academic research is consistent with their intended use.
We provide ATD-Trans for academic research to applicants who agree to the terms of use, in accordance with Japanese copyright law; the copyright of each travel blog post is retained by its respective travelogue writer.
As with the original ATD data, the source and translated text does not contain any information about the travelogue authors.


\paragraph{Human Workers}
The annotation work was performed by human annotators at an annotation company, and the translation work was performed by human translators managed by a translation company.
Payments to the companies were based on their submitted estimates.
The individual annotators and translators, as well as their compensation, were determined by the respective companies.
Under the contracts for the annotation and translation work, the intellectual property rights to the deliverables were transferred to the client, i.e., one of the authors or the authors' institution.

\section*{Acknowledgments}
This study was partially supported by JSPS KAKENHI Grant Number JP23K24904 for geo-link annotation of the overseas dataset.

\bibliography{custom}

\clearpage
\appendix

\section{Detailed Experimental Settings} \label{app:exp_settings}
\paragraph{List of Evaluated Models.}
We used the following LLM checkpoints available at Hugging Face for the MT experiments:
\vspace{-0.75pc}
\begin{itemize}
\setlength{\itemsep}{0pt}
\setlength{\parskip}{0pt}
\setlength{\parsep}{0pt}
{\small
\item \path{meta-llama/Llama-3.1-8b-Instruct},
\item \path{meta-llama/Llama-3.1-70b-Instruct},
\item \path{tokyotech-llm/Llama-3.1-Swallow-8B-Instruct-v0.3},
\item \path{tokyotech-llm/Llama-3.1-Swallow-70B-Instruct-v0.3},
\item \path{google/gemma-2-9b},
\item \path{google/gemma-2-27b},
\item \path{tokyotech-llm/Gemma-2-Llama-Swallow-9b-it-v0.1}, and
\item \path{tokyotech-llm/Gemma-2-Llama-Swallow-27b-it-v0.1}.
}
\end{itemize}

\paragraph{Prompts.} \label{app:prompt}
The prompts used in the MT experiments are listed in Table~\ref{tab:prompts}: \#1 for the paragraph-context prompt, \#2 for the document-context prompt, \#3 for the document-context prompt with KB-predicted or gold terms, \#4 for the document-context prompt with oracle terms, and \#5 for the document-context prompt for the ID-replaced data (\S\ref{sec:syn_ids}).
Notably, angle brackets <{}< >{}> appearing in model outputs are omitted during evaluation.

\paragraph{Metrics.}
The d-BLEU scores were computed with SacreBLEU~\cite{post-2018-call} ({\small \path{nrefs:1|case:mixed|eff:no|tok:intl|smooth:exp|version: 2.4.3}}).

\begin{table}[t!]
\small
\centering
\begin{tabular}{lp{6.2cm}}
\toprule
\#1 & Translate the following Japanese paragraph into English. Output only the translated paragraph.\\
& Japanese: \texttt{\{paragraph\}}\\
& English:\\
\midrule
\#2 & Read the following Japanese document (a total of \texttt{\{num\_segs\}} paragraphs, each beginning with "\ja{★}") for context only. Then translate the target paragraph into English. Output only the translation of the target paragraph.\\
& Full document context: \texttt{\{context\}}\\
& Target paragraph (Japanese): \texttt{\{paragraph\}}\\
& English translation:\\
\midrule
\#3 & Read the following Japanese document (a total of \texttt{\{num\_segs\}} paragraphs, each beginning with "\ja{★}") for context only. Then translate the target paragraph into English. \blue{In the translation, replace each <{}<location expression>{}> in the text with the appropriate English term using the provided mappings (format:  <{}<source term\,$||$\,target term>{}>). Use the mapped target term only if it is contextually appropriate. Otherwise, translate the location expression normally. In all cases, preserve the <{}< >{}> brackets around each translated term in your output.} Output only the translation of the target paragraph.\\
& Full document context: \texttt{\{context\}}\\
& Term mappings: \texttt{\{mapping\}}\\
& Target paragraph (Japanese): \texttt{\{paragraph\}}\\
& English translation:\\
\midrule
\#4 & Read the following Japanese document (a total of \texttt{\{num\_segs\}} paragraphs, each beginning with "\ja{★}") for context only. Then translate the target paragraph into English. \blue{In the translation, replace each <{}<location expression>{}> in the text with the appropriate English term using the provided mappings (format: <{}<source term\,$||$\,target term>{}>). Use the mapped target term only if the source term substantially corresponds to the location expression and the mapping is contextually appropriate. An exact textual match is not required. Otherwise, translate the location expression normally. In all cases, preserve the <{}< >{}> brackets around each translated term in your output.} Output only the translation of the target paragraph.\\
& Full document context: \texttt{\{context\}}\\
& Term mappings: \texttt{\{mapping\}}\\
& Target paragraph (Japanese): \texttt{\{paragraph\}}\\
& English translation:\\
\midrule
\#5 & Read the following Japanese document (a total of \texttt{\{num\_segs\}} paragraphs, each beginning with "\ja{★}") for context only. Then translate the target paragraph into English. \blue{Location and infrastructure names in the Japanese text are represented by ID strings (e.g., L001 for locations, F001 for facilities, T001 for public transport vehicles, and R001 for roads, waterways, or public transport lines). Keep these identifiers unchanged in the translation.} Output only the translation of the target paragraph.\\
& Full document context: \texttt{\{context\}}\\
& Target paragraph (Japanese): \texttt{\{paragraph\}}\\
& English translation:\\
\bottomrule
\end{tabular}
\caption{Prompts used in the experiments. Parts specific to each prompt is highlighted in \blue{blue}.}
\label{tab:prompts}
\end{table}

\begin{table*}[t]
\small
\centering
\begin{tabular}{lcc|ccc|ccc}
\toprule
\multirow{2}{*}{System} & \multirow{2}{*}{Context} & \multirow{2}{*}{Term} & \multicolumn{3}{c|}{Domestic data} & \multicolumn{3}{c}{Overseas data} \\
\cmidrule{4-9}
&&& \texttt{\!d-BLEU\!} & \texttt{COMET} & \texttt{\!TermAcc\!} & \texttt{\!d-BLEU\!} & \texttt{COMET} & \texttt{\!TermAcc\!} \\
\midrule
\multirow{4}{*}{gemma-2-9b} 
& Doc & None & 29.8 & 79.6 & 45.2 & 35.2 & 85.0 & 79.0 \\
& Doc & KB pred &\bpink 28.0 &\bpink 76.4 &\bblue 47.7 & -- & -- & -- \\
& Doc & KB gold &\bpink 28.1 &\bpink 76.5 &\bblue 47.7 &\bpink 32.1 &\bpink 82.7 &\bpink 74.9 \\
& Doc & Oracle &\bblue 33.8 &\bblue 81.3 &\bblue 99.2 & 34.9 & 84.1 &\bblue 99.1 \\
\midrule
\multirow{4}{*}{Gemma-2-Llama-Swallow-9b-it-v0.1\!\!\!} 
& Doc & None & 30.6 & 79.6 & 55.5 & 34.5 & 83.9 & 78.5 \\
& Doc & KB pred &\bpink 28.8 &\bpink 77.7 &\bpink 47.2 & -- & -- & -- \\
& Doc & KB gold &\bpink 28.7 &\bpink 77.6 &\bpink 46.8 &\bpink 31.6 &\bpink 82.8 &\bpink 76.2 \\
& Doc & Oracle &\bblue 33.7 &\bblue 82.4 &\bblue 98.3 & 34.6 & 83.5 &\bblue 98.5 \\
\midrule
\multirow{4}{*}{gemma-2-27b}
& Doc & None & 32.7 & 80.7 & 57.5 & 36.4 & 85.2 & 79.9 \\
& Doc & KB pred &\bpink 28.8 &\bpink 76.0 &\bpink 45.9 & -- & -- & -- \\
& Doc & KB gold &\bpink 28.6 &\bpink 76.1 &\bpink 44.9 &\bpink 32.3 &\bpink 81.6 &\bpink 70.1 \\
& Doc & Oracle &\bblue 34.9 & 81.5 &\bblue 98.7 & 35.6 &\bpink 83.8 &\bblue 98.8 \\
\midrule
\multirow{4}{*}{Gemma-2-Llama-Swallow-27b-it-v0.1\!\!\!}
& Doc & None & 33.2 & 81.2 & 59.3 & 36.6 & 85.0 & 81.1 \\
& Doc & KB pred &\bpink 32.0 &\bpink 79.8 &\bpink 57.2 & -- & -- & -- \\
& Doc & KB gold &\bpink 31.5 &\bpink 79.7 &\bpink 56.6 &\bpink 35.0 & 84.3 &\bpink 78.2 \\
& Doc & Oracle &\bblue 37.2 &\bblue 82.8 &\bblue 99.6 &\bblue 37.6 & 85.1 &\bblue 99.2 \\
\bottomrule
\end{tabular}
\caption{Translation quality of Gemma-2 and Gemma-2-Llama-Swallow with term-specifying prompts.} \label{tab:res_mt2_gemma}
\end{table*}

\section{Geocoding for MT Experiments} \label{app:geocoding}
To obtain KB-predicted terms in the experiments described in \S\ref{sec:term_prompt}, we used the BERT-based bi-encoder model implemented in the POI Geocoder tool~\cite{nakatani-etal-2025-text}, with default hyperparameters. The geocoding model, trained on the ATD-MCL training set of the official 7:1:2 data split, achieved a recall@1 of 0.55 and a recall@5 of 0.84 on the ATD-MCL test set.\footnote{Our test set for the domestic data is a subset of the ATD-MCL test set.}

The bilingual terms used in the KB term prompt were obtained as follows.
For each input gold geo-entity mention in a paragraph, the geocoder predicted an entry group, which is a set of OSM entries that refer to almost the same real-world locations (the preprocessed DB with this grouping is available in the POI Geocoder repository).
After entry groups were predicted, within each entry group, we sorted node, way, and relation entries separately by their ID strings in ascending order and selected the first three entries of each type, resulting in at most nine entries in total. 
From each selected entry, we obtained a (Japanese name, English name) pair via the Nominatim lookup API\footnote{\scriptsize \url{https://nominatim.org/release-docs/develop/api/Lookup/}}
 when both attributes were available, and used the most frequent pair within each entry group. Entries whose English name field contained Japanese strings were excluded.

\begin{table}[t!]
\centering
\includegraphics[width=7.7cm]{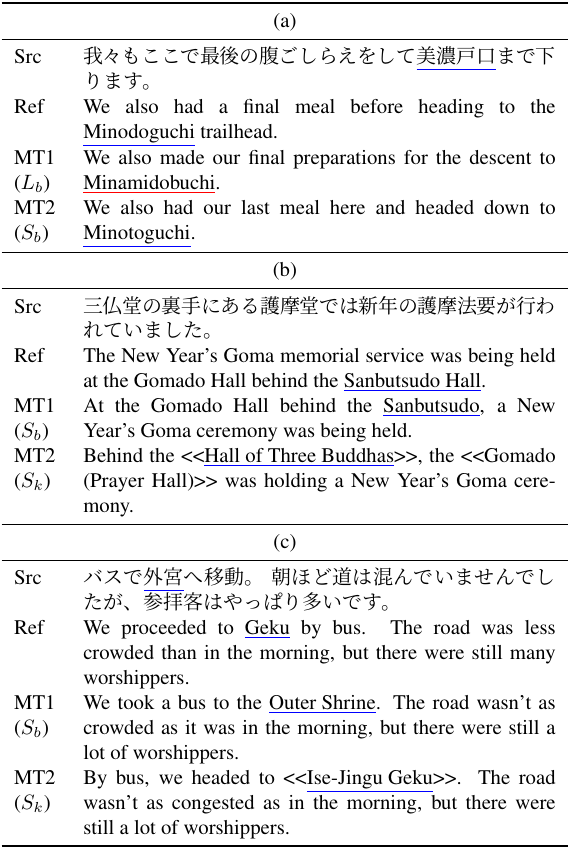}
\caption{Examples of source texts (Src), reference translations (Ref), and model outputs (MT\{1,2\}) for the dev sentences from the domestic data. $L_b$ denotes the Llama-3.1 70B model with the basic document-context prompt. $S_b$ and $S_k$ denote the Llama-3.1-Swallow 70B model with the basic document-context prompt and KB gold term prompt, respectively.}
\label{fig:mt_example}
\end{table}

\section{Additional Experimental Results}
\subsection{MT Results of Gemma-based Models} \label{app:gemma}
Table~\ref{tab:res_mt2_gemma} shows the results for the Gemma-2 and Gemma-2-Llama-Swallow models with term-specifying prompts.
Similarly to the results for the Llama-3.1 and Llama-3.1-Swallow models reported in \S\ref{sec:term_prompt}, we observed the following trends: (i) the oracle term prompt improved almost all metrics, with Term Accuracy reaching nearly 100\%, and (ii) the KB gold and KB pred term prompts decreased d-BLEU and COMET scores compared with the basic prompt.

\subsection{Translation Examples} \label{app:mt_example}
Table~\ref{fig:mt_example} shows example outputs from selected models.\footnote{In Example (a), \textit{Minotoguchi} appears to be the more common reading of ``\ja{美濃戸口}'' and is used in the local bus stop name, but we also found several English web articles that refer to it as \textit{Minodoguchi}, similar to the reference term.}
Detailed analyses are provided in \S\ref{sec:mt_example}.
The KB gold term prompts for Examples (b) and (c) contain the specifications ``<{}<\ja{三仏堂}\,$||$\,Hall of Three Buddhas>{}>'' and <{}<\ja{伊勢神宮 外宮}\,$||$\,Ise-Jingu Geku>{}>,'' respectively.

\subsection{Geo-Entity Mention Recognition} \label{app:mr}
Geo-entity mention recognition is a fundamental task in geographic text analysis.
We evaluated the accuracy of existing systems on geo-entity mention recognition, formulated as a named entity recognition (NER) problem targeting location names (\texttt{LOC\_NAME}) and facility names (\texttt{FAC\_NAME}), using both the source Japanese text and the translated English text, each of which includes domestic and overseas data.
The total number of annotated mentions is shown in Table~\ref{tab:data_split}.
We report span-level F1 scores based on exact match with gold spans, without considering label type, thereby focusing purely on mention boundary accuracy.

\begin{table}[t]
\small
\centering
\begin{tabular}{p{7.25cm}}
\toprule
\ja{入力文から、地名・地形名（ラベル：LOCATION}\\
\ja{）および施設名（ラベル：FACILITY）に該当する固有表現をすべて抽出し、"固有表現:ラベル;固有表現:ラベル;..."の形式で、出現順に列挙してください。該当するものがなければ"NONE"のみ出力してください。}\\
\ja{入力：}\texttt{\{sentence\}}\\
\ja{出力：}\\
\midrule
From the input sentence, extract all named entities that correspond to place names and geographical feature names (label: LOCATION) and facility names (label: FACILITY), and list them in the order of appearance in the format "named entity:label; named entity:label; ...". If no such entities are found, output only "NONE".\\
Input: \texttt{\{sentence\}}\\
Output:\\
\bottomrule
\end{tabular}
\caption{Prompt used in the mention recognition experiments and its English translation (10-shot examples are abbreviated).}
\label{tab:prompts_me}
\end{table}

\subsubsection*{Systems}
As off-the-shelf NER tools based on masked language model (MLM), we evaluated spaCy (en\_core\_web\_trf 3.0.0)~\cite{spacy} and GiNZA (ja\_ginza\_bert\_large\_$\beta$1)~\cite{ginza-2019}.
We also fine-tuned a multilingual LUKE model (mluke-large-lite)\footnote{\scriptsize \url{https://huggingface.co/studio-ousia/mluke-large-lite}}~\cite{ri-etal-2022-mluke} on the ATD-MCL training set using the LUKE-NER span extraction framework.\footnote{\url{https://github.com/naist-nlp/luke-ner}}

For LLMs, we evaluated Llama 3.1 (8B, 70B) and Llama-3.1-Swallow (8B, 70B) models\footnote{We regard causal language models with 1B or more parameters as LLMs. We used the same Llama models as those employed in the MT experiments and Instruct-v0.1 models for the Llama-3.1-Swallow series.} using 10-shot in-context learning with fixed 10 sentences, following the prompt shown in Table~\ref{tab:prompts_me}.

\begin{table}[t]
\small
\centering
\begin{tabular}{l|ccc|ccc}
\toprule
& \footnotesize{P} & \footnotesize{R} & \footnotesize{F1} & \footnotesize{P} & \footnotesize{R} & \footnotesize{F1} \\ 
\midrule
\midrule
System & \multicolumn{3}{c|}{Ja-Domestic} & \multicolumn{3}{c}{Ja-Oveseas} \\
\midrule
GiNZA & 63.3 & 73.7 & 68.1 & 71.9 & 75.6 & 73.7 \\
mLUKE${}^\dagger$ & \underline{87.8} & \underline{78.8} & \underline{82.8} & \underline{84.1} & \underline{84.0} & \underline{84.1} \\
\midrule
Llama${}^\text{8B}$    & 33.1 & 35.3 & 34.2 & 37.3 & 19.0 & 25.1 \\
Swallow${}^\text{8B}$  & 39.8 & 83.5 & 53.9 & 48.6 & 78.6 & 60.1 \\
Llama${}^\text{70B}$   & 61.9 & 38.1 & 47.2 & \underline{78.0} & 51.2 & 61.8 \\
Swallow${}^\text{70B}$\!\! & \underline{69.2} & \underline{86.7} & \underline{77.0} & 714 & \underline{81.3} & \underline{76.1} \\
\midrule
\midrule
System & \multicolumn{3}{c|}{En-Domestic} & \multicolumn{3}{c}{En-Oveseas} \\
\midrule
spaCy & 75.3 & 73.0 & 74.1 & 77.3 & 80.4 & 78.8 \\
mLUKE${}^\dagger$ & \underline{87.6} & \underline{84.9} & \underline{86.2} & \underline{80.8} & \underline{88.1} & \underline{84.3} \\
\midrule
Llama${}^\text{8B}$    & 71.7 & 72.3 & 72.0 & 74.3 & 68.9 & 71.5 \\
Swallow${}^\text{8B}$  & 52.7 & 80.9 & 63.9 & 64.0 & 73.1 & 68.3 \\
Llama${}^\text{70B}$   & \underline{86.6} & 72.3 & 78.8 & \underline{81.3} & 69.8 & 75.1 \\
Swallow${}^\text{70B}$\!\! & 76.4 & \underline{83.5} & \underline{79.8} & 76.1 & \underline{74.9} & \underline{75.5}\\
\bottomrule
\end{tabular}
\caption{Precision (P), recall (R), and F1 score of mention recognition systems on the ATD-MCL test set (0--100 scale). Underlined values indicate the best score within each block. For fine-tuned models ($\dagger$), the reported scores are averaged over three runs.} \label{tab:res_mr}
\end{table}

\subsection*{Results and Discussion}
Table~\ref{tab:res_mr} reports the results. Fine-tuned mLUKE achieved the highest accuracy across all four datasets, with F1 scores around 85. 
No clear accuracy differences were observed across languages and regions.
Among LLMs, Swallow-70B achieved relatively high accuracy (F1 $\geq$ 75) on both Japanese and English data, despite using only 10-shot in-context learning. This model also showed no substantial regional accuracy gap.

A notable observation is that Llama performed poorly on Japanese data, while achieving strong accuracy on English data in both domestic and overseas regions. This suggests that Llama has limited capability in recognizing Japanese-script geo-entity names while its knowledge of Japanese location and facility names in English form is comparatively robust.

\subsection{Geocoding} \label{app:ed}
Another fundamental task in geographic text analysis is geocoding. 
We evaluated the accuracy of existing systems on geocoding, formulated as a mention-level entity disambiguation task.
We adopt a four-choice question format that is feasible even for computationally intensive LLMs.\footnote{For the Japanese domestic data, under the full geocoding setting in which entries are predicted from the entire database, the accuracy of the BERT-based geocoder is as reported in \ref{app:geocoding}.
Conducting a similar experiment on the overseas data would require preprocessing the large raw OSM database by aggregating entries at the entry-group level for POIs outside Japan. Moreover, evaluating LLMs under this setting would necessitate constructing a pipeline that combines a candidate retrieval method with the use of LLMs for candidate reranking.}
Specifically, given a sentence containing a target mention, the model is requested to select the correct OSM entry from four candidates. 
The distractor candidates consist of entries with similar names or randomly selected entries. For both the domestic and overseas datasets, we restrict the evaluation to mentions with valid gold OSM entries and further limit the cases to those referring to locations inside Japan for the domestic data and outside Japan for the overseas data. The resulting number of questions is reported in Table~\ref{tab:num_menqa}.

\begin{table}[t]
\footnotesize
\centering
\begin{tabular}{ccc|ccc}
\toprule
\multicolumn{3}{c|}{Domestic} & \multicolumn{3}{c}{Overseas} \\
train & dev & test & train & dev & test \\
\midrule
415 & 165 & 322 & 465 & 169 & 289 \\
\bottomrule
\end{tabular}
\caption{Number of questions in the geocoding data.}
\label{tab:num_menqa}
\end{table}

\begin{table}[t]
\small
\centering
\begin{tabular}{p{7.25cm}}
\toprule
\ja{入力テキスト中の<{}<場所参照表現>{}>が指している場所を、与えられた選択肢（1,2,3,4）の中から選んでください。}\\
\ja{入力：以前から<{}<出雲>{}>に行きたいと思っていましたが、中国山地を列車で越えて見たいとの思いもあり、岡山経由で山陰に入ることにしました。}\\
\ja{選択肢：}\\
\ja{1: name="出雲市". full\_name="出雲市, 島根県, 日本". category=boundary. type=administrative. addresstype=city.}\\
\ja{2: name="出雲". full\_name="出雲, 穂波嘉穂線, 大字九郎丸, 飯塚市, 嘉穂郡, 福岡県, 日本". category=highway. type=traffic\_signals. addresstype=highway.}\\
\ja{3: name="出雲". full\_name="出雲, 亀岡園部線, 千歳町千歳, 亀岡市, 京都府, 621-0001, 日本". category=highway. type=bus\_stop. addresstype=highway.}\\
\ja{4: name="出雲". full\_name="出雲, 国道165号, 黒崎, 桜井市, 奈良県, 633-0017, 日本". category=highway. type=traffic\_signals. addresstype=highway.}\\
\ja{回答：}\\
\midrule
Select the location referred to by the <{}<location reference expression>{}> in the input text from the given options (1, 2, 3, 4).\\
Input: I had wanted to visit <{}<Izumo>{}> for a long time, and I had also wanted to cross the Ch\=ugoku Mountains by train, so I decided to enter the San'in region via Okayama.
Options: (omitted)\\
Answer:\\
\bottomrule
\end{tabular}
\caption{Prompt used in the geocoding experiments and its English translation, with input and option placeholders instantiated using a concrete example (4-shot examples are abbreviated).}
\label{tab:prompts_ed}
\end{table}

\begin{table}[t]
\small
\centering
\begin{tabular}{l|cccc}
\toprule
System & Ja-Dom & Ja-Ovs & En-Dom & En-Ovs \\
\midrule
Exact Match & 30.5 & 23.7 & 28.2 & 40.1 \\
Fuzzy Match & \underline{51.3} & \underline{46.7} & \underline{48.4} & \underline{50.5} \\
\midrule
RoBERTa-J  & \underline{53.4} & \underline{53.0} & 19.3 & 17.2 \\
XLM-R      & 33.9 & 23.2 & \underline{28.6} & \underline{29.8} \\
\midrule
Llama${}^\text{8B}$     & 74.2 & 83.9 & 82.3 & 81.4 \\
Swallow${}^\text{8B}$   & 75.5 & 84.6 & 78.9 & 79.7 \\ 
Llama${}^\text{70B}$    & 79.2 & 88.4 & \underline{84.5} & 83.2 \\
Swallow${}^\text{70B}$  & \underline{80.8} & \underline{88.8} & 83.2 & \underline{83.5} \\
\bottomrule
\end{tabular}
\caption{Accuracy of geocoding systems on the ATD-MCL test set (0--100 scale; Dom: domestic data, Ovs: overseas data). Underlined values indicate the best score within each block.} \label{tab:res_ed}
\end{table}

\subsubsection*{Systems}
As rule-based baselines, we evaluated Exact Match (EM) and Fuzzy Match (FM). EM and FM retrieve candidate entries based on exact string match or surface similarity, respectively, and randomly select one from the candidates.

For MLM-based approaches, we evaluated a Japanese RoBERTa Large model (denoted RoBERTa-J)\footnote{\scriptsize \url{https://huggingface.co/nlp-waseda/roberta-large-japanese}}~\cite{liu2020roberta} and XLM-R Large\footnote{\scriptsize \url{https://huggingface.co/FacebookAI/xlm-roberta-large}}~\cite{conneau-etal-2020-unsupervised}. The input sentence and each candidate option were encoded into vector representations; the candidate with the highest inner product similarity to the sentence representation was selected.

For LLMs, we used the same four models as in \S\ref{app:mr}. Using the prompt in Table~\ref{tab:prompts_ed}, we conducted 4-shot in-context learning with fixed four questions and selected the answer with the highest log-likelihood among choice tokens (``1''--``4'').

\subsubsection*{Results and Discussion}
Table~\ref{tab:res_ed} reports the results. FM achieved approximately 50\% accuracy across datasets. Among MLM-based methods, RoBERTa-J outperformed FM only on Japanese data.
All four LLMs achieved generally high accuracy. Contrary to our expectations, Llama achieved performance comparable to or better than Swallow of the same size, even on Japanese data. No clear performance differences across languages were observed among LLMs.
One possible explanation is that the multiple-choice format reduces the need for extensive prior knowledge of Japanese place names; the task may be solvable to a certain degree even with limited geographic knowledge.

\end{document}